# THE ROLE OF TIME IN THE CREATION OF KNOWLEDGE

Roy E. Murphy, Ph.D.
3042 Carpenter Loop SE
Olympia, WA 98503
360.491.5227

RoyMurphy@Computer.org

**ABSTRACT**

In this paper I assume that in humans the creation of knowledge depends on a discrete time, or stage, sequential decision-making process subjected to a stochastic, information transmitting environment. For each time-stage, this environment randomly transmits Shannon type information-packets to the decision-maker, who examines each of them for *relevancy* and then determines his optimal choices. Using this set of relevant information-packets, the decision-maker *adapts,* over time, to the stochastic nature of his environment, and optimizes the *subjective expected* rate-of-growth of knowledge. The decision-maker's optimal actions, lead to a decision function that involves, over time, his view of the subjective entropy of the environmental process and other important parameters at each time-stage of the process. Using this model of human behavior, one could create psychometric experiments using computer simulation and real decision-makers, to play programmed games to measure the resulting human performance.

Scientists seek to understand the experience of our environment. Some build hypothetical, mathematical models that reflect our view of reality as they adumbrate the *laws of nature,* enabling them to conduct experiments leading to the validation of a hypothesis as they reach out for even more truths about nature. A scientific model builder must steer a narrow path between the naivety of oversimplification and the morass of over complication. "*Knowledge Creation"* by human decision-makers, over a sequence of time based environmental *events,* is the basis of this paper. Knowledge creation systems, often named "learning systems," have had a long and successful history in psychometrics. The pioneering works of R. R. Bush and F. Mosteller (1955) led to the development of sequential, time based mathematical structures that simulate the creation of knowledge by experiments in lower animal learning. The voluminous works of P. Suppes, for example, (1956), extended these structures to human learning experiments. The influence of knowledge creating systems that effect human decision-making has resulted in the application of several mathematical discrete time-sequential system analysis techniques that stem from the classical works of H. Poincaré and G.D.Birkhoff (1927), and lead on to the recent works of the late Richard Bellman (1957) of the RAND Corporation. These "dynamic programming" techniques have also been successfully modeling time-sequential scientific processes. Applying these same mathematical techniques to the analysis of the creation of knowledge over sequential-time in *human* decision-makers may appear to be a "tail wagging the dog" to some, but new ideas come by strange paths. The subject of what time and human knowledge *is* has occupied philosophers from Plato to Husserl. It has now become possible to analyze knowledge creation as a discrete, time-sequential stochastic, adaptive decision-making process, subjected to the laws of probability, and the step-by-step actions of humans, who in Kenneth Arrow's words, are *learning by doing* (K. Arrow, 1961).

In this paper, in its simplest form, I will develop a mathematical structure of a human knowledge creation over time as a mathematical process of this type. Despite the simplicity of this model, the mathematical details are very extensive and, for the sake of the balance between the simple and the complex, I have avoided many formulations that can be found in R. Murphy, (1965).

Observation of a human *decision-maker* over time, behaving in a knowledge creating process, indicates that, on the average, his intelligence increases over time which tends to simplify the decision-maker's process by decreasing his *relative entropy*, a measure of his personal uncertainty stemming from his view of his environment. This process occurs in the face of a stochastic *environment*, a "deus ex machina," that feeds the decision-maker a variety of random *information-packets,* as time dependent events. Some of these, time dependent, information-packets are valuable to the decision-maker, but most are irrelevant. Somehow, the decision-maker is able, for each time event, to sort these random information packets into a list of increasing *acceptance* from the irrelevant to the relevant, depending on the decision-maker's previous experience, stored in his *memory*. In this process, the decision-maker tends to *decrease*, his relative entropy (his personal uncertainty), in contradistinction to the chaotic world of non intelligent dynamic systems where the net entropy is believed to be always increasing. In the Brownian motion of micro-particles in suspension, for example, it appears that these non-intelligent systems lack a means for *organizing against*, or *adapting to* the hidden mechanisms of their environment. For the most part, except for some chemical processes that tend oscillate back and forth over time from chaos to order, with obvious consequences to the effect on their entropy, these non-intelligent processes, in the end, always *increase* their local entropy. Ilya Prigogine (1996) has given many examples of this kind of pseudo lifelike behavior.

In this paper, I postulate that lifelike systems have several *necessary* unique facilities that affect their decrease in relative entropy through the experience of time-events during their organized knowledge creating activity. The first of these is the basic ability to make a *sequence* of optimal decisions over time. This ability, in the face of a very complex world, enables living beings to *adapt* and *control* their decision-making through the *acceptance or rejection* of random information-packets that are transmitted from their environment during each time-event. Within these systems, we find the human knowledge creation activity, over time, enables the optimal acceptance or rejection of random information-packets, received by their senses from their immediate stochastic environment and then stored in their mind. As Shakespeare aptly put it for poets (and mathematicians),

> *"And as imagination bodies forth*
> *The forms of things unknown, the poet's pen*
> *Turns them to shapes, and gives to airy nothing*
> *A local habitation and a name."*
>                    "A Midsummer Night's Dream,"
>                    Act V, Scene I.

In general, human knowledge creation systems must possess a unique and necessary *memory* capability that enables them to optimize their personal and social decision-making behavior. The research of the late Hans Reichenbach (1999) in his book

*The Direction of Time* has shown that memory (of some kind) is an *essential* element forcing time to always seem to flow in an irreversibly *increasing* direction. This human memory, the second *necessary* element, enables the human decision-maker, to *reduce* his relative entropy and *optimize* his decisions over time. The mathematical model, described herein, considers how these key dependences, decision-making and memory, enable the human knowledge creating process and produce a decrease in the decision-maker's relative entropy.

It takes nerve for an economist and computer scientist to leap into a world dominated by cognitive scientists and philosophers, and worse yet, propose yet another mathematical model for man's capability to accumulate knowledge. My previous research in mathematical statistics, information theory and economic decision processes forced me to pass down this dangerous path. It began when Claude Shannon (1948) introduced a rational measure for information based on the needs of the communication engineers at Bell Telephone Labs and thus launched the formal substructure for "Information Theory." In his original concept, Shannon considered that *all* information-packets had *equal* value. Since his primary focus was on the efficiency of the *transmission* of electronic information over a bandwidth limited communications channel, he was not directly concerned with the *value* of that information. In Shannon's original view the *value* of a transmitted information-packet was always in the eyes of the *sender* or *receiver,* not the communications service provider. Leon Brillouin (1963) pointed out that the *value* of Shannon's measure of information is dependent on the potential *use* of that information by a decision-maker over time. Shannon and his associates soon rectified that omission in their later papers. J. Marshack (1960) advanced the classical theory of economics by considering the role of information as a new "factor" in the production function. His work was soon followed by several other similar papers. Recently, Kenneth Arrow (1996) extended Shannon's cost of information theory by showing how information theory, applied to modern economic theory, causes the generation of strong increasing economic returns to scale, over time, in production that dominates today's new economic growth in an "information" age. Finally for non-mathematical philosophers, Fred Dretske (1999) has recently distilled, in great detail, all of these knowledge and information fragments. These new economic and philosophic applications of information theory provide a jumping off point for this paper.

In the early 1900's, Henri Poincaré gave a lecture before the Societe de Psychologie in Paris on Discovery in Mathematics. This lecture forms chapter III of his book, *Science and Method*. Poincaré (1953) , agreeing with Helmholtz, felt that the phenomenon of knowledge creation was composed of three distinct time phases. After attending Poincaré's famous lecture, Jacques Hadamand (1954) pulled all this together in his book, *The Psychology of Invention in the Mathematical Field.* He named the three distinct time phases as, "Preparation, Incubation and Illumination." I will paraphrase these three time-phase names in less formal terms:

1. Research it!
2. Sleep on it!
3. Aha!

Poincaré explained his view of these time phases:

> *"One is at once struck by these appearances of sudden illumination, obvious indications of a long course of previous unconscious work. The part* played *by this unconscious work in mathematical discovery seems to me indisputable, and we shall find traces of it in other cases where it is less evident. Often when a man is working at a difficult question, he accomplishes nothing the first time he sets to work. Then he takes more of less of a rest, and then sits down again at his table. During the first half hour he still finds nothing, and then all at once the decisive idea presents itself to his mind."*

Hadamand's concept of "Incubation," or as I call it, the "Sleep on it" time phase, is a familiar experience for most scientists as they "invent" new concepts. The "Aha" phase always seems to appear later, almost any time *after* the "Incubation" time-phase or as I call it the "Sleep on it" time phase. I will postulate that each time-stage in my knowledge creation process passes through each of these three time phases. Poincaré goes further and borrows Epicurus's interesting "idea hooks" argument and gathers a call for wide diversity in research. He says:

> *"Among the combinations [of ideas] we choose, the most fruitful are often those which are formed of elements borrowed from widely separated domains."*

I might also add to Poincaré's concept of Epicurean idea hooks, a reference to Leibniz's concept of monadology where mysterious supernatural beings transmit ideas to humans! More realistically, a fascinating recent study on the "Sleep On It" phase by Matthew Walker (2000), "Sleep to Remember," describes in detail the beneficial effect of multiple cycles of REM (Rapid Eye Movement) sleep on developing human memory. Many psychometric and neuropsychological studies are reported in the literature about the importance of REM sleep on the creation of ideas. It has been said that Einstein, when asked how much sleep he needed to create physical theory, impishly replied – at the least, ten hours a night!

Now, I must take a broader view of information, not only the *value* of information but also the role of information as a *creator* of knowledge. Consider a TV viewer watching a cable news broadcast, over time, a broad range of Shannon information-packets assail him, some of them are intensely important to him, others are possibly interesting and still others are virtuously useless. From experience we know that the TV watcher is able to *classify* the information-packets he receives, by some measure of *value*, and thus make decisions based on these value judgments.

I propose to define a measure, or as Bellman (1960) would have it, a "control" vector, that I will call the decision-maker's *information-relevance* vector. Each component of this vector is the relevance the decision-maker attaches to each of the types of information-packets he receives from his environment during a time-event. Using this information-relevance concept, consider the evaluation of the $i^{th}$ information increment the decision-maker receives during the $t^{th}$ stage or time-event in a discrete sequential decision process. We can postulate a possible value for this relevancy for each component as

$$k_{i,t} = r_{i,t} y_i \tag{1}$$

where

$r_{i,t}$ is the $i^{th}$ component of the decision-maker's information-relevance vector assigned to the $i^{th}$ kind of information-packet during time-stage $t$ of his discrete sequential decision process,

$y_i$ is the $i^{th}$ component of an independent random valued vector that measures the *Shannon* information content of an information-packet for the $i^{th}$ *kind* of information, and

$k_{i,t}$ is the $i^{th}$ component, also an independent random valued vector, and is a measure of the decision-maker's *relevant* information from the $i^{th}$ information-packet, transmitted to him by the environment, during the $t^{th}$ time-stage of his sequential decision process.

So how can the decision-maker compute the *optimal* value ($r_{i,t}$) for each information-packet from the $i^{th}$ source during stage $t$? Since the information–packets are components of a random vector, determined solely by the decision-maker's environment, I propose that the decision-maker will compute the optimal value for each $r_{i,t}$ based on the maximization of the statistical *expectation* of the compounded rate-of-growth of knowledge from each of the $i^{th}$ information-packet sources during each previous time stage of the process. Because of this "rate-of-growth" evaluation, the statistical expected value of the *compounded* growth of knowledge for the process takes on a natural logarithmic form.

Why a compounded rate of growth? Why a logarithmic function? The most familiar example of compounded rate-of-growth is in the conventional evaluation of an investment for one's retirement fund. Suppose each year you reinvest the earned increment of your fund, so that the value of your fund will grow exponentially as time passes. So also we can view the decision-maker's investment in his knowledge, since his knowledge is a kind of investment for his life. One could postulate, "the more you know today, the more you are *capable* of knowing tomorrow;" or "what the decision-maker will know tomorrow is what you know today *plus* the contribution of any new *relevant* information you have gained today." In other words, a student must know about the concept of *numbers* to be able to know about the concept of *algebra* and he must know about *algebra* to be able to know about the concept of *calculus*.

Let's take a look at a very simple discrete *deterministic* compounded sequential knowledge creating process, over time, for the compounded rate-of-growth of knowledge, $K_T$, given a constant increment of a decision-maker's relevant information given during each time-stage of the process. We form the following deterministic discrete sequential process

$$K_1 = K_0(1+k),$$
$$K_2 = K_1(1+k),$$
$$\cdots$$
$$K_T = K_{T-1}(1+k).$$

By repeated substitution, this leads to a more familiar, concise, result

$$K_T = K_0(1+k)^T,$$

taking the natural logarithm of this we get

$$g(K_0) \triangleq \frac{1}{T}\ln\frac{K_T}{K_0} = \ln(1+k), \tag{2}$$

where

$g(K_0)$ is defined as the *compounded rate-of-growth of knowledge* for a *T* stage process.

## The Decision-Maker's "Research It" Phase Begins

In the decision-maker's "Research it" time phase, I postulate a mathematical model for the decision-maker's compounded rate of growth of knowledge by considering a simple *single stage* process at time-stage *t*

$$\ln\frac{K_{t+1}}{K_t} = \sum_{i=1}^{M} n_{i,t} \ln(1+r_{i,t}y_i) \quad i=1,2,\ldots M \tag{3}$$

where

M is the number of components in the random valued environmental vector, that is, it is the *size* of the set of information-packet types the environment can transmit,

$i$ is the index for the $i^{th}$ component or kind of information–packet,

$t$ is the event-time index signifying the $t^{th}$ stage of the process, and

$n_{i,t}$ is the *frequency* (the number of ) transmissions for the $i^{th}$ kind of information-packet during time-stage $t$ of the process.

Remembering that the $y_i$'s are components of an independently chosen *random* vector and since that forces $K_{t+1}$ to also be a random vector, it is customary to view this process *analytically* by considering the *statistical expectation* of this random process. So, we have for *one* typical time-stage of the decision-maker's knowledge process currently at time-stage $t$

$$E\left\{\ln\frac{K_{t+1}}{K_t}\right\} = \sum_{i=1}^{M} E(n_{i,t})\ln(1+r_{i,t}y_i), \text{ for } i=1,2,...M. \quad (4)$$

Now it is also realistic to assume that the probability distribution for the environment's random value vector, the information generator, is to be chosen from a multinomial beta distribution. Therefore, in this special case, the statistical expectation of $n_{i,t}$ is given by

$$E(n_{i,t}) = t\, p_i \quad (5)$$

where we also insist on the usual probability space conditions,

$$p_i \leq 1 \text{ and } \sum_{i=1}^{M} p_i = 1, \; i=1,2,...M, \quad (6)$$

and we also assume that this random process is quasi-ergodic – that is the probabilities, the $p_i$'s, will remain constant during the time duration of the process.

So in this case, it can be shown that the rate-of-growth for our time-sequential knowledge creation process at time-stage t is given by

$$\bar{g}_t \triangleq E\left\{\frac{1}{t}\ln\frac{K_t}{K_{t-1}}\right\} = \sum_{i=1}^{M} p_i \ln(1+r_{i,t}y_i), \text{ for } i=1,2,...M, \quad (7)$$

where

the "$\frac{1}{t}$" signifies that this is for just *one* stage of a process after already experiencing *t* time-stages.

The information-relevance vector can be subjected to the additional special conditions

$$r_{i,t} \geq 0 \text{ and } \sum_{i=1}^{M} r_{i,t} = 1, \text{ for stage } t \text{ for } i = 1, 2, ... M, \tag{8}$$

and each of the environment's information-packages is assigned to a probability of $p_i$ during time-stage *t*.

At this point equation (7) and the constraints of (6) and (8) completes the decision-maker's "Research it" phase for a single current $t^{th}$ stage.

## The Decision-maker's "Sleep-On-It" Phase Begins

The *optimal* expected compounded rate-of-growth of knowledge for a single stage *t* of a *T* stage process can now be evaluated. This operation begins a decision-maker's "Sleep on it" phase, where he mulls over during his "REM" sleep or in his subconscious, the importance of the information-packets he has received during his "Research it" phase and decides what to do about it. In the "Sleep on it" phase, I postulate that the decision-maker constructs in his mind an ordered list for each component of the information-relevance vector he received, the set of $r_{i,t}$'s, that tends to maximize the stage's rate of growth. Because of constraints, in (6) and (8), on the $p_i$'s and the $r_{i,t}$'s, finding the optimal values for the decision-maker's, the $r_{i,t}$*'s, at stage *t* is, for example, a *non-linear* mathematical optimization procedure. The simple "marginal value" procedure for optimizing equation (7), one that is based only on the determination of the first and second derivatives of equation (7), will not assure that the special constraints on the $p_i$'s and the $r_{i,t}$'s will hold.

Negative components of the decision-maker's information-relevance vector, the $r_{i,t}$'s, are questionable. Since I have assumed that *zero* relevance is already an automatic *rejection* of an information packet, as far as an actual contribution to the decision-maker's optimal rate-of-growth of knowledge, for these rejected packets, nothing really happens! On the other hand, a negative $r_{i,t}$, could be considered some form of conscious unknowing or forgetting for that kind of information-packet. This expansion of the model seems mathematically overly complex at this point, and I will not consider it in this paper.

A prototype for a non-linear optimization technique that will enable a decision-maker to create this intuitive list of optimal *relevant* values for the $r_{i,t}$'s for each kind of information-packet he received form the environment at time-stage *t* is needed. This mathematical model should emulate all that could possibly go on in the decision-maker's subconscious while he decides to *accept or reject* each information-packet delivered by his environment. There is such a non-linear optimization technique for determining a list of constrained optimal $r_{i,t}$'s that could be used to generate this list of relevant components under the non-linear constraints. This optimization technique is an expanded version of the classical method of Lagrange, coupled with special conditions that handle the non-linear constraints as set forth by the Kuhn-Tucker (1951) theorem. Other more "computer" oriented methods such as the "Branch and Bound" procedure, Wagner (1969), could also be used, but I will employ the typical Kuhn-Tucker procedure here. This method divides the indexes of the $y_i$'s, into two subsets, $i = 1, 2, \ldots, M^+$, that is of size $M^+$, for the first set and $i = (+1), \ldots, M$ for the second set that is of size $M^0$. It is assumed that the first subset of the $y_i$'s, will meet the special constraints for $r_{i,t} \in M^+$ and the second subset of $y_i$'s will violates the special constraints for $r_{i,t} \in M^0$.

However, in reducing the components of the set of *i*'s we have altered the probability space upon which the $p_i$'s were originally defined. In other words, it is possible that $\sum_{j=1}^{M^+} p_j < 1$. Therefore, we must *renormalize* the original probability space to that represented only by the set of $i = 1, 2, \ldots, M^+$ where $M^+ \leq M$. This can be done with the relations

$$q_i = \frac{p_i}{\sum_{j=1}^{M^+} p_j}, \qquad (9)$$

$$q_i \leq 1, \text{ and } \sum_{i=1}^{M^+} q_i = 1.$$

The set $\{q_i\}$ now reflects the proper probability space over just the $M^+$ subset assigned to the information-packets that the decision-maker has determined that are *relevant* for the optimization of $\bar{g}_t$. Assuming that this procedure is now employed, we get the equation for determining the optimal relevant $r_{i,t}$'s

$$\bar{g}_t^* \triangleq \max_{r_{i,t}} \left\{ E \ln \frac{K_t}{K_{t-1}} \right\} = \max_{r_{i,t}} E \sum_{i=1}^{M^+} \left[ q_i \ln\left(1 + r_{i,t} y_i\right) \right]. \quad i = 1, 2, \ldots, M^+. \qquad (10)$$

By using the references, and using the Kuhn-Tucker the non-linear optimization method, we can find the $r_{i,t}^*$'s that optimize (10) is given by

$$r_{i,t}^* = q_i \left[ 1 + \sum_{j=1}^{M^+} \frac{1}{y_j} \right] - \frac{1}{y_i}, \qquad i = 1, 2, \ldots, M^+, \qquad (11)$$

where

$M^+ \in M$ is the sub-set of the decision-maker's relevant information-packets.

## The Decision-Maker Adapts to his Environment

The decision-maker's "Sleep on it" phase needs further consideration. There is a serious human problem with equations (9), (10) and (11). The decision-maker really *does not know* the values of the $p_i$'s and thus also does not know the $q_i$'s at time-stage *t*! What the decision-maker *does know,* is in his *memory* that results from his history of the time-past *frequency of occurrences* of the types of information-packets previously transmitted by the environment. In other words, the frequency of occurrence of the $y_i$'s. I now assume that the decision-maker will *adapt* to his stochastic environment by using this mental history to *estimate* the value if these unknown $q_{i,t}$'s and thereby enabling him to compute the *adaptive* optimal $r_{i,t}$'s in an equation similar to (11).

Note, that now the estimated or *subjective* values of the probabilities are subscripted with an addition of a "*t*" because these estimated values of the $q_{i,t}$'s will be developed by adapting *over time* to the history of the environment over a number of time-stages, say for example all *t* time-stages of a *T* stage process. An appropriate statistical method to enable the decision-maker to adapt to the environment is to assume that the unknown probabilities are defined by the multinomial beta distribution. Given this assumption, the decision-maker can determine or *estimate* the adaptive or subjective $\hat{q}_{i,t}$'s (the hat "^" indicates that the $\hat{q}_{i,t}$ are the *subjective* probabilities) by the Bayes estimation procedure, Good (1965). The *actual* $q_i$'s, of course, are known only to the "deus ex machina" (God's random environment). There are many other statistical procedures that closely resemble this possible mental or subjective process, but I propose the method of Bayes, for several reasons. This is an *intuitive* procedure that would be natural for a human decision-maker to mentally or subjectively undertake. The Bayes estimator also has the useful property for being statistically *consistent* and *sufficient*.

This is one major reason to consider the Bayes estimator as rational procedure because with it the estimation can begin as a complete *guess* on the part of the decision-maker and still evolve, over time, into an approximation of the actual probabilities. Suppose initially the decision-maker has *no* time-history of any past occurrences of the $y_i$'s, that is, he is at the initial stage of the process. Carnap (1950) has suggested an initial (a priori) equation, as a "hunch" or "guess," that can "start" the decision-makers learning experience to estimate the $\hat{q}_{i,t}$'s over time. Using his method we have then for the first stage of the process, where the decision-maker guesses the values for the subjective probabilities for the occurrence of each kind of information-packet

$$\hat{q}_{i,0} = \frac{\alpha_i}{\sum_{j=1}^{M^+} \alpha_j} \ , \ i, j = 1, 2, \ldots, M^+, \tag{12}$$

where the $\alpha_i$'s are Carnap's "logical width" parameters and represent the decision-makers *initial* guesses for the $\hat{q}_{i,t}$'s.

Now if the decision-maker has progressed as far as time-stage *t,* he should possesses a history of the actual environmental events over his *t* stages, he would have for the subjective estimate of the probabilities

$$\hat{q}_{i,t} = \frac{n_{i,t} + \alpha_i}{t + \sum_{i=1}^{M^+} \alpha_j}, \quad i, j = 1, 2, \ldots, M^+, \tag{13}$$

where the $n_{i,t}$'s are the decision-maker's observations of the frequency of occurrence for each of the random $y_i$'s, up to and including the $t^{th}$ stage.

One particular set of values for Carnap's parameters for an initial a priori estimate of the probabilities has important properties. If the decision-maker has *no* guess for the initial priorities it would be appropriate to postulate the Laplace assumption, that initially the probability for *every* information-packet is as likely as any other. In other words the set $\{\alpha_i\}$ has one identical value for $\alpha_i$ in (12) and (13) for *all* $i = 1, 2, \ldots, M^+$. In this case, such a decision-maker will be suffering from (Laplacian) *ignorance*. In total, given this type of ignorance, the a priori value of the probabilities would be simply be

$$\hat{q}_{i,0} = \frac{1}{M^+}, \text{ for } i = 1, 2, \ldots, M^+, \text{ for time-stage 0.} \tag{14}$$

We will assume for simplicity that the decision-maker will start in total ignorance of these initial probabilities. Note that after a few time-stages of the process and if $n_{i,t}$ observations for the information-packets have occurred, the $\hat{q}_{i,t}$'s are very likely to deviate from this equal likely condition. On the first few time-stages of this *adaptive*, stochastic knowledge creating process, when the decision-maker still has only a few values for the $n_{i,t}$'s, the decision-maker may exhibit wild instability (partial ignorance) in his estimates for the $\hat{q}_{i,t}$'s. Fortunately, since the Bayes estimator is known to be both statistically consistent and sufficient, the estimator is one that will "settle down" over time and even have an "end" in time, that is, if the estimation of the $\hat{q}_{i,t}$'s continues forever it will ultimately converge, in probability, to the true $q_i$'s. Thus we can say

$$q_i = \plim_{t \to \infty} (\hat{q}_{i,t}) \text{ for } i = 1, 2, \cdots, M^+.$$

We can now write the *adaptive* version of equations (10) and (11) that reflect the decision-maker's use of *subjective* probabilities to generate the optimal adaptive relevance factors, the $\hat{r}_{i,t}$'s, by using the decision-maker's now *known* subjective probabilities. We have

$$\overline{g}_t^* \triangleq \max_{r_{i,t}} \left\{ \hat{E} \left[ \frac{1}{t} \ln \frac{K_t}{K_{t-1}} \right] \right\} = \max_{r_{i,t}} \hat{E} \sum_{i=1}^{M^+} \left[ \hat{q}_{i,t} \ln \left(1 + r_{i,t} y_i \right) \right], \qquad (15)$$

and

$$\hat{r}_{i,t}^* = \hat{q}_{i,t} \left[ 1 + \sum_{j=1}^{M^+} \frac{1}{y_j} \right] - \frac{1}{y_i}, \quad \text{for } i, j = 1, 2, \ldots, M^+, \qquad (16)$$

also

$$\hat{q}_{i,t} = \frac{\hat{p}_{i,t}}{\sum_{j=1}^{M^+} \hat{p}_{i,j}}, \quad i, j \in M^+. \qquad (17)$$

Remember that $\sum_{j=1}^{M^+} \hat{p}_{i,j}$ *may* not be equal to 1, since $M^+ \leq M$.

By substitution of (16) into (15) we can determine the optimal rate-of-growth of knowledge for this single $t^{th}$ stage. This substitution will bring up several important relationships with respect to the entropy of the environmental process, but first we must consider the *whole time-event sequential process* in detail. With the completion of the equations (15) (16) and (17) above we have come to the end of the "Sleep on it" phase for *one stage*, the $t^{th}$, of a discrete, sequential, *adaptive,* stochastic knowledge-gathering process, we now are ready to turn to the "Aha" phase, where we finalize the whole sequence of T stages, where $t = 1, 2, \ldots, T$.

## The "Aha!" Phase Begins

We have seen above that the adaptive estimates of the subjective probabilities require some kind of history of the decision-maker's observations, so we must extend the concept developed above to an *entire* event-time sequence, where for each stage, *t*, the decision-maker's *relevant* sequential event history can be generated.

First, it is important that the basic mathematical structure for multistage time-event sequential processes of this type be well understood before we continue and apply this technique to the knowledge creating process. A sequential discrete time process is composed of a series of *linked*, causally ordered (irreversible) *time-stages*. Each stage begins and ends with one or more similar values, or vectors, called the *input state* and produces the *output state*. During the execution of a time-stage, one or more *events* occur that transform the input state into the output state. The input state is *transformed* into the output state by the time-events of the current stage. In our *stochastic* version of such a process, one or more of the state input vectors are independently randomly distributed vectors. In these cases, the stage's time-event transformation involves taking the *statistical expectation* of the random state variables over all the historical time-events. This use of the expectation function makes it possible to analyze mathematically, stochastic processes of this type.

The backbone of many of these discrete sequential processes is based on Markov's concept, see for example Doob (1952), for stage-by-stage transformations of expectation values of these random state variables. In general, for a Markovian discrete sequential process, after any number of time-stages, say *t,* we insist that the value of the remaining ($T$-$t$) stages will depend *only* on the values of the state variables at the end of the $t^{th}$ stage. Thus, in a Markovian process, all the historical content that the decision-maker has gathered during the process up to the $t^{th}$ stage is transformed and carried *forward* to the next $(t+1)^{th}$ stage in the form of the current output state vector. This means that for a Markovian process we can be assured that the decision-maker will not have *lost* any relevant history by using the previous state vector at stage *t*, that might be required for the decision-maker in the following stages.

It is important to be clear that, because of assuming Markov properties, the decision-maker does not need to *remember* the results of the entire sequence of the history of his past events to determine his new, current subjective probabilities. He needs only to remember his last subjective probabilities to re-compute his newest subjective probabilities from the previous values by *transforming* the former Bayes estimates into the *new* Bayes estimates using the new frequency data he obtains because of the occurrence of this new environmental time-event. An objection voiced by many neurological scientists, regarding the massive number of neurons required for the human brain to store a complete historical record, is avoided, at least in this mathematical model, thanks to the recursive nature of the Bayes estimation procedure and the Markovian property for sequential processes. Therefore, this estimation derived from the decision-maker's past history, encapsulates all the result of the previous events into a single, mental current vector. This single current mental vector could be defined as the decision-maker's current *conviction,* generated by the history of all the previous results during his decision process.

The reliance on the Markovian property and Bayes statistical estimation procedure in this mathematical model is of sufficient importance to warrant review of the classical problem of "memory verses chronological record" so clearly outlined by Heinz Von Foerster (1967). If the number of neurons, *n*, in the human brain were counted (*n* is estimated as $10^{10}$) then the possible ways of interconnecting between all these neurons would be $\Omega = 2^{n^2}$ possible ways. This is a very large number of possible networks or ways to connect these neurons. Using this estimate, one can calculate the total number of information bits that can be stored in any one network of neurons we call the human brain. Since there is no reason to assume one network is more likely than another, it can be estimated as $H = \ln 2^{n^2} = n^2$ bits of information per neuron network. Using an $\Omega$ as above, the brain could store $H = (10^{10})^2 = 10^{20}$ bits per brain. This is even a larger number! Yet it still is probably insufficient to provide a *chronological* record of all the event-outcomes experienced during the lifetime for a human decision-maker. An estimate for the number of bits of information in a *single* zygote of a human genetic program has been estimated as $10^5 < H < 10^{12}$. So clearly, the brain, however complex, could not be capable of keeping a chronological record for all the event-outcomes experienced by a typical decision-maker.

In this paper, I am suggesting that some method of encapsulating, or folding this huge number of chronological records into the limited *memory* capacity of the human brain is necessary to make rational human decisions. Mathematically speaking this method is to encapsulate the decision-maker's event-outcome record, by assuming the Markovian property, so that the current level of information at the current time-stage is dependent *only* on the immediately proceeding level of information. In other words, at each time-stage, all the *relevant* information that the decision-maker has experienced during his past time-stages has been encapsulated or folded into the immediate preceding level of information and that, and that alone, becomes the initial conditions for the processing of the current time-stage. In this stochastic time-stage process this information encapsulation is accomplished by assuming that the process is Markovian and employs a Bayesian (or other) statistically *sufficient* estimator.

For example: We *know* that kicking on big stones is hard on your foot. Do we really know this or are we just *convinced* that we know this? Some decision-makers, who have inborn common sense would accept this opinion. They are already convinced that kicking a stone *must* hurt your foot, so *conviction* plays a sort of initial condition in the knowledge-gathering process. For the rest of us, how many times must we kick a big stone before we are convinced that it hurts? Do you need to remember *every* time you kicked a big stone? Probably not, you too will have accumulated all the knowledge you need to be convinced about big stones. The existence of your *conviction* is the analog of the Bayesian statistical estimator in our mathematical process for knowledge-gathering. If you, as a decision-maker perceive in the current time-stage the presence of a stone in your path, chances are you decision would be to step past it and go on to your next time-stage. Did you review in your mind *all* the past events when you tripped on stones in your past time-stages? Chances are you didn't, because you possessed the *conviction* that big stones in your path should be avoided and that *single* conviction actually occupied a relatively small number of the available information bits stored in your brain. *Conviction* becomes the economizer of our brain information bits, and plays the role of the sufficient statistic in our mathematical model. Without conviction, our memory would soon run out of sufficient bits to enable us to make rational decisions.

With the use of the details we have developed about sequential dynamic processes we can now see how the decision-maker can maximize an *entire* time-sequential knowledge creating process consisting of *T* stages. This extension is not just a simple substitution of *T* and *T-1* for the single stage *t* and *t-1* equations. Fortunately, maximizing the *first* stage of the time sequential process is relatively easy, since the decision-maker has, by definition, no previous time history on which to base estimates of the subjective probabilities (and therefore he relies on his Carnap *guess* for the initial subjective probabilities) and only his initial conditions, his initial state of knowledge, $K_0$. As we saw in equations (15) and (16), the decision-maker has sufficient information to determine the optimal values of his relevancy vector for the *first* stage of the *T* stage process, but just how *does* the decision-maker determine this optimal values for all the rest of the stages of the process?

Writing the equation for an entire *T* stage optimal process in an explicit format we get,

$$\hat{g}*(K_0) = \frac{1}{T} \max \hat{E}\left[\ln\frac{K_T}{K_{T-1}} + \ln\frac{K_{T-1}}{K_{T-2}} + \cdots + \ln\frac{K_2}{K_1} + \ln\frac{K_1}{K_0}\right]. \tag{18}$$

The " * " here indicates that the $\hat{g}*(K_0)$ is its "optimal" value.

Because of the additive property of statistical expectations we can rewrite (18) as,

$$\hat{g}*(K_0) =$$
$$\frac{1}{T}\left[\max \hat{E}\ln\frac{K_T}{K_{T-1}} + \max \hat{E}\ln\frac{K_{T-1}}{K_{T-2}} + \cdots + \hat{E}\max\ln\frac{K_2}{K_1} + \hat{E}\max\ln\frac{K_1}{K_0}\right]. \tag{19}$$

Because *each* of the above terms is based on the immediately previous value of the rate-of-growth of knowledge and the previous values of the subjective probabilities for the information-packets, except the first term that is based on the initial condition $K_0$ and the initial guess of the subjective probabilities (that the decision-maker already knows) we can maximize each of these terms with respect to each time-stage's information-relevance vector, one at a time, recursively. We have here a multistage Markov transformation equation where each element has an unknown input to produce an unknown output, except for the first stage, that the decision-maker already *knows*. We can formulate this function as a "dynamic program" as introduced by Richard Bellman. We define the Bellman functional relationship for the one, *first* stage process, given, $K_0$, and Carnap's, $\hat{q}_{i,0}$. We have

$$\hat{f}_1(K_0) \triangleq \max_{\{r_{i,0}\}} \hat{E}_0 \ln K_1 = \max_{\{r_{i,0}\}} \sum_{i=1}^{M^+} \hat{q}_{i,0} \ln(1 + r_{i,0} y_i) + \ln K_0 \qquad (20)$$

and

$\hat{q}_{i,0} = \dfrac{\alpha_i}{\sum_{i=1}^{M^+} \alpha_i}$ where the $\alpha_i$'s are Carnap's "logical width", that is, the decision-maker's initial guesses of the estimators for the initial subjective probabilities. Continuing along Bellman's same functional scheme, we have for the second and following stages

$$\hat{f}_2(K_0) \triangleq \max_{\{r_{i,1}\}} \sum_{i=1}^{M^+} \hat{q}_{i,1} \ln(1 + r_{i,1} y_i) + \hat{f}_1(K_0) \qquad (21)$$

$$\bullet \ \bullet \ \bullet$$

$$\hat{f}_T(K_0) \triangleq \max_{\{r_{i,T-1}\}} \sum_{i=1}^{M^+} \hat{q}_{i,T-1} \ln(1 + r_{i,T-1} y_i) + \hat{f}_{T-1}(K_0) \qquad (22)$$

Following Bellman's approach to the solution of dynamic processes, equation (22) encloses the entire sequence of terms with in the brackets of equation (19) to one functional equation. Returning to the objective of the paper, the maximum compounded rate-of-return for knowledge creation, we form the final result

$$\hat{g}_T *(K_0) = \frac{1}{T}\left[\hat{f}_T(K_0)\right] \qquad (23)$$

This terse functional is best analyzed using computer simulation.

## Implications of the Results

Referring back to equations (15) and (16), it is appropriate to actually substitute the decision maker's optimal relevance vector from (16) into (15). For example, we can maximize a typical $t^{th}$ stage of the compounded rate-of-growth for knowledge creation and obtain some interesting single stage results. After considerable algebra, we get the *optimal* functional for a *typical*, $t^{th}$ stage as

$$f_{t+1}(K_0) = (H_t^* - \hat{H}_t) + \hat{E}_t(\ln Y_t) + \hat{E}_t\left\{\ln\left[\frac{1}{M^+} + \frac{1}{\Phi}\right]\right\} + f_t(K_0) \qquad (24)$$

where

$f_t(K_0)$ is the *optimal* value (already determined by the decision-maker) from the previous t-1 time-stages of the process,

$$\hat{E}_t(\ln Y_t) = \sum_{i=1}^{M^+} \hat{q}_{i,t} \ln y_i ,$$

$\hat{H}_t \triangleq -\sum_{i=1}^{M^+} \hat{q}_{i,t} \ln \hat{q}_{i,t}$ (this is the decision-maker's subjective entropy for the environment at the current time-stage), and since $\sum_{i=1}^{M^+} \hat{q}_{i,t} = 1$,

$H_t^* \triangleq \sum_{i=1}^{M^+} \hat{q}_{i,t} \ln \frac{1}{M^+} = -\ln M^+$ (this is the *maximum* subjective entropy for the environmental process at the current stage).

Furthermore, $\Phi$ is the *geometric mean* for all the information-packets in set $\{Y_t\}$ for $i \in M^+$, where $\Phi$ is defined by,

$$\Phi \triangleq \frac{M^+}{\sum_{j=1}^{M^+}\frac{1}{y_i}}.$$

The notation, $\hat{E}_t$ always means that we are taking the expectation with respect to the decision-maker's set of current subjective probabilities, $\{\hat{q}_{i,t}\}$, at each time-stage $t$.

So then what is the entropy, $\hat{H}_t$, mean? According to Shannon,

*"H is a measure of how much "choice" is involved in the selection of the event or how uncertain we are of the outcome."*

In other words, $\hat{H}_t$ is the *decision-maker's* subjective measure of the *uncertainty* about the underlying stochastic environmental information-packet generation process at stage $t$.

Now the maximum of $H_t$ is $H_t^*$, and it occurs when *all* the possible kinds information-packets at the current stage, appear to be *equal likely,* (the Laplace assumption) *as viewed by the decision-maker,* thus there is in this case a *maximum* degree of uncertainty for the decision-maker's choice. So he is indifferent, and he may chooses to receive *all* or *none* of them. In other words, the decision-maker would be in the state of total ignorance! On the other hand, if any one information-packet is *for certain*, that is, its probability is unity, the rest must be zero (or not transmitted) so there is *no* uncertainty about the optimal relevance-vector for the certain information-packet, and therefore $\hat{H}_t$ is zero. (Note: 1 ln 1 is taken as 0). In this case the decision-maker just accepts only the single relevant information-packet and passes on to the next stage.

What then is the quantity $(H_t^* - \hat{H}_t)$? It is often considered as a measure of marginal or *relative* uncertainty. We note that Shannon has defined

$$C \triangleq H^* - H \tag{25}$$

as a system's "information channel capacity." The information capacity is zero if the probabilities defining $H$ (that are generating the randomness) are equal. On the other hand the information capacity is unity if any one of these probabilities is one and therefore $H$ is zero.

On the other hand, Heinz Von Foerster (1967) has suggested a similar function that he has called a measure of "order" of a stochastic process. Von Foerster's *order* function is given as

$$\Omega \triangleq \frac{H^* - H}{H^*} = 1 - \frac{H}{H^*}. \tag{26}$$

A random process is in a completely *disorganized* state is when $H = H^*$, the condition of total ignorance, where anything and everything is possible, that is, all probabilities are independent and equal; thus $\Omega = 0$. On the other hand, if the process is deterministic and thus in perfect order, $H$ is zero and $\Omega = 1$.

Now consider a time-sequential decision process for the knowledge creating process, where the decision-maker estimates the subjective probabilities based on his experiencing continuing reception of information-packets as he goes along during each time-stage of the transformational process. We note that there is an *actual* entropy for the knowledge creating process that, by our definition, is independently generated by the random process transmitting the information-packets to the decision-maker. We have assumed that the decision-maker has no control over this environmental process and, therefore, has no control over the magnitude of the actual $H$ of the process.

Now consider our time-sequential decision-making process for knowledge creation where the decision-maker estimates the *subjective* probabilities based on his reception of random information-packets during each stage. In the flavor of Shannon's definition of channel capacity, I propose a subjective measure of a decision-maker's *relative entropy* or uncertainty for an adaptive time-sequential decision process given by

$$\hat{\theta}_t \triangleq |H - \hat{H}_t|, \tag{27}$$

where the actual entropy of the process, $H = -\sum_{i=1}^{M} p_i \ln p_i$, is fixed and based on the probabilities of the *entire* set of $\{y_i\}$, $i = 1, 2, ..., M$. If the actual entropy of the environmental process is $H$, no matter what the decision-maker's initial estimate of his subjective, $\hat{H}_t$, might be, then even though he might start totally ignorant, his relative entropy would fall over time. We have

$$\hat{H}_t \underset{t \to \infty}{\to} H, \text{ and } \hat{\theta}_t \underset{t \to \infty}{\to} = 0,$$

and because the Bayes estimation procedure is *consistent*, the decision-maker's relative entropy for the knowledge creation process, in probability, would tend toward zero as time extends to infinity.

## Some Concluding Remarks

This hypothetical model assumes the ability of a human *decision-maker* to draw out of his memory of the past events and adapt to the most rewarding information-packets available to him, while rejecting all the others, thereby, enabling him to optimize his compounded rate-of-growth of knowledge. At the risk of over simplifying an impossibly difficult human process, I hypothesize that:

*Intelligent decision-makers attempt to maximize their subjective expectation of their compounded rate-of- growth of knowledge by partitioning, at each time-stage, a set of received information-packets into two sets; one of acceptable, and another of unacceptable packets. The decision-maker does this selection by mentally (using his memory) determining an optimal **relevancy-vector**, and that provides a means for him to divide the information-packets into these two subsets. Acting on these relevant information-packets the decision-maker chooses an optimal set that, in probability, **decreases** his relative entropy during his sequence of experiential events as he maximizes his expected compounded-rate-of growth of knowledge over the time duration of his decision-making activities.*

Many simplifying technical assumptions must be made to make this hypothesis amenable to mathematical analysis. Some of these assumptions are:

1. The process can be decomposed into a sequence of discrete causal *time-stages*, each of which receives the *a posteriori state* from the previous stage. During each time-stage a decision-maker takes an *action* that generates the *a priori state* for the following time-stage.
2. During this transformation the decision-maker optimizes his statistical expected compounded rate-of-growth of knowledge at each time-stage of the process. This assumption is based on the simple concept, "The more you know today; the more you are *capable* of knowing tomorrow."
3. The environment for this process is characterized by a hidden underlying stochastic mechanism that produces a discrete sequence of mutually independent *events,* the information-packets transmitted to the decision-maker and generated by a given quasi-stationary statistical process of the multinomial beta type.
4. The *events* produced by the environment at each stage of the discrete sequential process for each type of information-packet are the transmission of

    Shannon information-packets that are made available to the senses of the decision-maker. At each stage of this process, the environmental stochastic process exclusively controls the type of this information-packet transmitted and the actual probability of this transmission is unknown to the decision-maker.

5. The decision-maker can only *observe* the stochastic events produced at each time-event by the environment, but not the underling probability generating parameters. This observation of events is augmented by the decision-maker's time dependent memory and that is used to construct a Bayes estimate at each stage for the subjective probability for each relevant event, given the arbitrary a priori distribution function of the multinomial beta type. The Bayes estimator, is statistically sufficient and consistent, and is used by the decision-maker to compute a *relevance-vector* for the information-packet types, at each time-stage in the process.

6. The decision-maker, during a time-stage of the process, is capable of determining a set of optimal, relevant information-packets that maximizes the subjective value of his compounded rate-of-growth of knowledge.

7. In the process of knowledge creation, the statistical expectation of the stochastic vectors in the process must be "Markovian," that is, the *current* statistical expectation can be estimated *solely* by the single immediately proceeding probability estimate and the environmental events that occur at the current time-stage. This means that the immediately proceeding statistical expectation is defined as the decision-maker's current *conviction* and must, in itself, contain a summation of all the estimation history of all the previous statistical expectations over the relevant time of the process. This assumption makes the decision-making process an *adaptive* process that can be optimized stage-by-stage, recursively.

8. Because Bayesian probability estimators are consistent, the subjective compounded rate-of- growth of knowledge approaches the theoretical rate-of-growth, in probability, as the number of time-stages in the process approaches time infinity. Thus an intelligent decision-maker gets more intelligent as the number of observed time-events increase, and the decision-maker's relative entropy, as defined in (27) above decreases as he adapts to his environment over the course of the time sequential process.

None of these assumptions are, for the most part, restrictive for a real knowledge creating process; but, as a final warning to all scientists, we quote the advice from the Buddha,

*Believe nothing, no matter where you read it,*
*or who said it,*
*no matter if I have said it,*
*unless it agrees with your own reason and*
*your own common sense.*

## Suggestions for Further Research

Testing the hypothesis presented in this paper is required to justify the assumptions made herein. The art of psychometric testing has entered a new technological phase. With modern personal computer technology and specialized computer game development software the psychometric researcher can create realistic decision-making situations with complex, but controlled, environments. Built-in, behind the scenes, random event generators, with probability driving functions known only to the experimenter, can easily be programmed. For example, presenting students with realistic decision-making computer games, coupled with behind the scenes data collection software can accumulation their reaction times and performance measures. Properly designed by careful experimenters, these techniques may verify hypotheses for human behavior, such as presented in this paper.

Finally, from the neuropsychological point of view, reverse memory searching, as the decision-maker relies on his memory to adapt and form estimates of subjective probabilities is in agreement with recent psychometric experiments with rats conducted by Davis J. Foster and Matthew A Wilson (2006) at MIT. These researchers were able to actually observe the "replay" of the rat's experienced memories by *measuring* the activity of the neurons in the rat's hippocampus region of the brain where *current* memory events are formed. Dr. Wilson seems to believe that the hippocampus region replays, in *reverse,* the rat's previous memorized events, then another part of the rat's brain, perhaps the prefrontal cortex, provides reward signals that enable a decision-function to determine which memory events are to be *retained* and which are to be *discarded* to generate an *advantageous* memory for the rat.